\documentclass[a4paper,twoside]{article}

\usepackage{epsfig}
\usepackage{subcaption}
\usepackage{calc}
\usepackage{amssymb}
\usepackage{amstext}
\usepackage{amsmath}
\usepackage{amsthm}
\usepackage{multicol}
\usepackage{pslatex}
\usepackage{apalike}
\usepackage{algorithm2e}
\usepackage[bottom]{footmisc}
\usepackage{SCITEPRESS}     

\usepackage{multirow}
\usepackage{version}
\usepackage{colortbl}
\usepackage{booktabs}
\usepackage{xcolor}
\usepackage{graphicx}
\newcommand{\wrt}{with respect to~}

\graphicspath{figures/}
\definecolor{gray}{rgb}{0.5,0.5,0.5}
\definecolor{orange}{rgb}{1.0,0.5,0}
\definecolor{green}{rgb}{0.1,0.8,0.1}
\definecolor{brown}{rgb}{1.0,0.,1.0}
\definecolor{black}{rgb}{0.0,0.0,0}
\definecolor{red}{rgb}{1.0,0.1,0.1}
\definecolor{blue}{rgb}{0.1,0.1,1.0}

\newcommand{\blue}[1]{\textcolor{blue}{#1}}

\newcommand{\black}[1]{\textcolor{black}{#1}}
\newcommand{\eqcolor}[1]{\textcolor{gray}{#1}}

\newcommand{\rg}[1]{\blue{RG: #1}}

\newcommand{\adf}{SDF}
\newcommand{\func}{FCO}
\newcommand{\poda}{SAM}

\def\m#1{\ensuremath{\mathtt{#1}}}

\def\v#1{\ensuremath{\mathbf{#1}}}
\def\mW{\m W}
\def\mI{\m I}
\def\mF{\m F}

\def\mH{\m H}
\def\mC{\m C}
\def\vx{\v x}
\def\vy{\v y}
\def\vp{\v p}
\def\vh{\v h}
\def\vm{\v m}

\def\normtwo#1{\left\lVert#1\right\rVert_{2}}
\def\parr#1{\left(#1\right)}
\def\curl#1{\left\{#1\right\}}
\def\croch#1{\left[#1\right]}

\def\tr{^{\top}}

\def\ap{MA}
\def\apkp{MA$_\text{text}$}

\begin{document}

\title{Are Semi-Dense Detector-Free Methods Good \\at Matching Local Features?}

\author{\authorname{
Matthieu Vilain\sup{1},
R{\'e}mi Giraud\sup{1},
Hugo Germain,
and Guillaume Bourmaud\sup{1}
}
\affiliation{
\sup{1}Univ. Bordeaux, CNRS, Bordeaux INP, IMS, UMR 5218, F-33400 Talence, France
}
\email{
\{matthieu.vilain, remi.giraud, guillaume.bourmaud\}@u-bordeaux.fr}
}


\abstract{Semi-dense detector-free approaches (\adf), such as LoFTR, are currently among the most popular image matching methods.
While \adf{} methods are trained to establish correspondences between two images, their performances are almost exclusively evaluated using relative pose estimation metrics. Thus, the link between their ability to establish correspondences and the quality of the resulting estimated pose has thus far received little attention.
This paper is a first attempt to study this link. We start with proposing a novel structured attention-based image matching architecture (\poda). It allows us to show a counter-intuitive result on two datasets (MegaDepth and HPatches): on the one hand \poda{} either outperforms or is on par with \adf{} methods in terms of pose/homography estimation metrics, but on the other hand \adf{} approaches are significantly better than \poda{} in terms of matching accuracy. We then propose to limit the computation of the matching accuracy to textured regions, and show that in this case \poda{} often surpasses \adf{} methods. Our findings highlight a strong correlation between the ability to establish  accurate correspondences in textured regions and the accuracy of the resulting estimated pose/homography.
Our code will be made available.}

\onecolumn \maketitle \normalsize \setcounter{footnote}{0} \vfill

\begin{figure}
\centering
\rotatebox{90}{\scriptsize \hspace{0.2cm} LoFTR+QuadTree \hspace{1.1cm} \poda{} (ours) }
   \includegraphics[width=.92\linewidth]{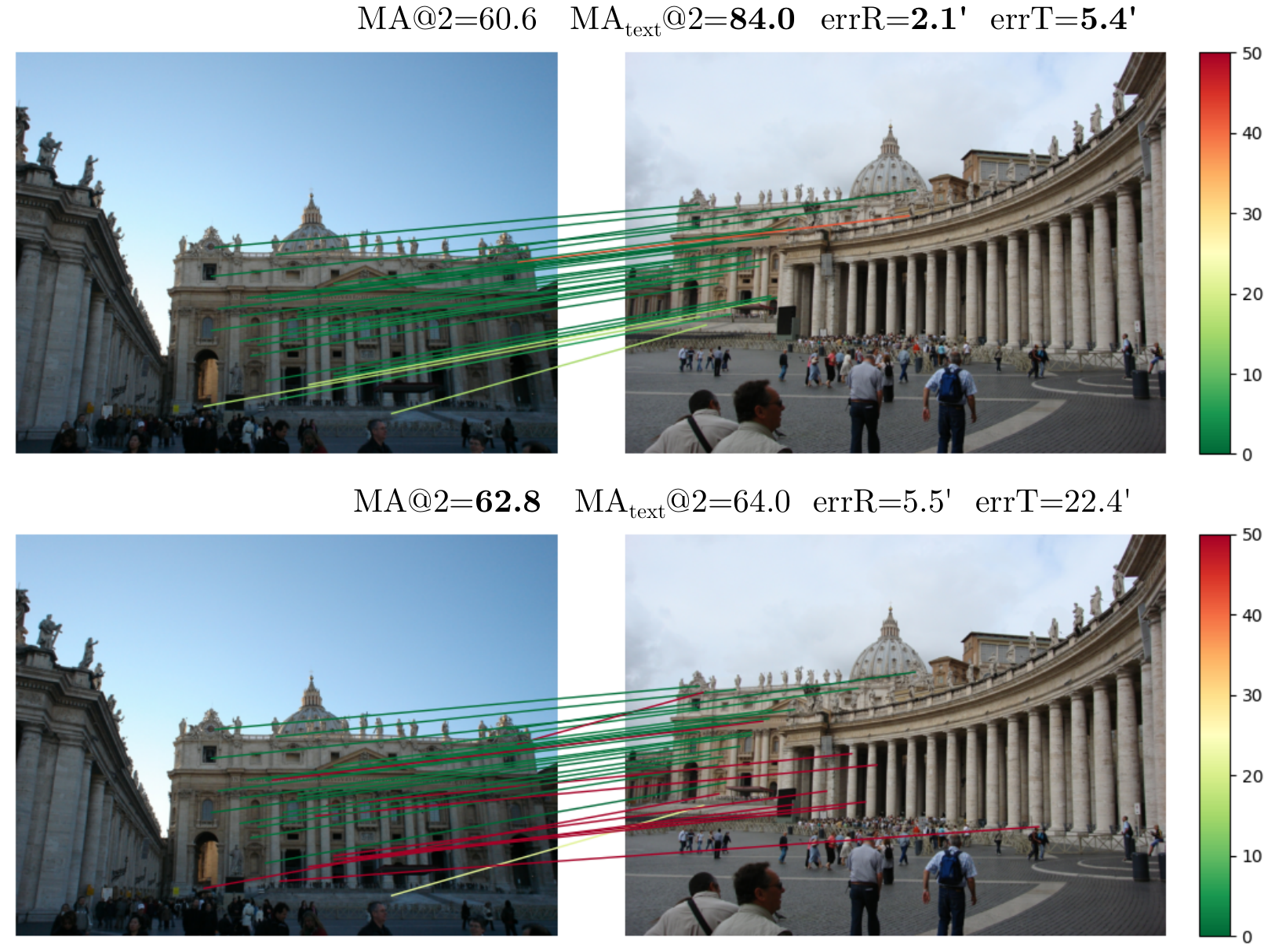}
   \caption{
   Given query locations within textured regions of the source image (left), we show their predicted correspondents in the target image (right) for: (\textbf{Top row}) \poda{} (Proposed) - Structured Attention-based image Matching,  (\textbf{Bottom row}) LoFTR~\cite{sun2021loftr}+QuadTree~\cite{tang2022quadtree} - a semi-dense detector-free approach  (line colors indicate the distance in pixels \wrt the ground truth correspondent).
   We report: (\ap@2) - the matching accuracy at 2 pixels computed on all the semi-dense locations of the source image with available ground truth correspondent (which includes both textured and uniform regions), (\apkp@2) - the matching accuracy at 2 pixels computed on all the textured  semi-dense locations of the source image with available ground truth correspondent (\emph{i.e.}, uniform regions are ignored), (errR and errT) - the relative pose error. \poda{} has a better pose estimation but a lower matching accuracy (\ap@2), which seems counter-intuitive. However, if we consider only textured regions (\apkp@2), then \poda{} outperforms
   LoFTR
   +QuadTree.} 
\label{fig:teaser}
\end{figure}

\section{Introduction}\label{sec:intro}

\begin{figure*}
  {\scriptsize
    \begin{tabular}{@{\hspace{0mm}}c@{\hspace{5mm}}c@{\hspace{0mm}}}
    \includegraphics[width=0.58\textwidth,height=0.275\textwidth]{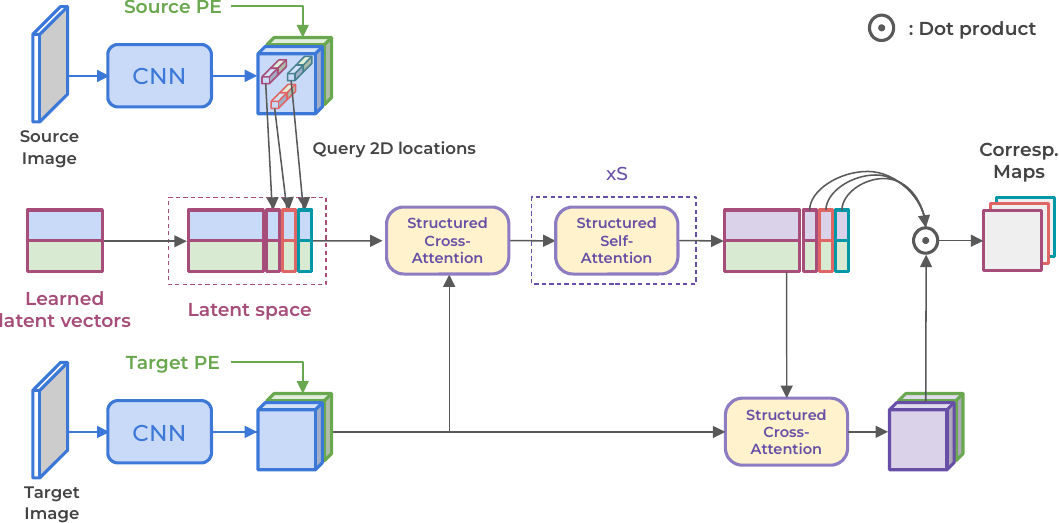} & 
    \includegraphics[width=0.39\textwidth]{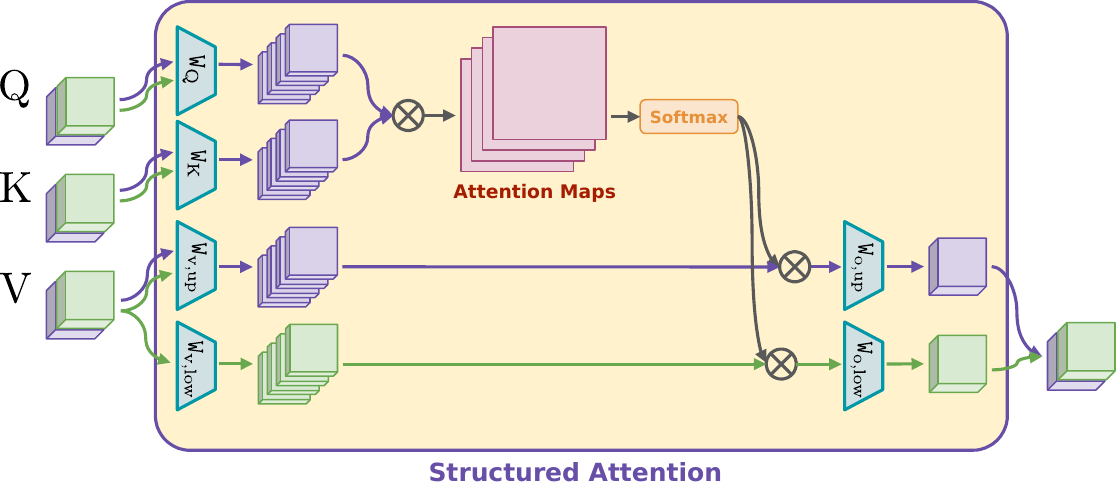} \\
    (a) \poda{} architecture& 
    (b) {Structured Attention}\\
  \end{tabular}
  }
  \caption{
  \textbf{Overview of the proposed Structured Attention-based image Matching (\poda{}) method.} 
  (a) The matching architecture first extracts features from both source and target images at resolution 1/4. Then it uses a set of learned latent vectors alongside the descriptors of the query locations, and performs an \emph{input} structured cross-attention with the dense features of the target. The latent space is then processed through a succession of structured self-attention layers. An \emph{output} structured cross-attention is applied to update the target features with the information from the latent space. Finally, the correspondence maps are obtained using a dot product. 
  (b) Proposed structured attention layer. See text for details.
  }
  \label{fig:baseline}
\end{figure*}

Image matching is the task of establishing correspondences between two partially overlapping images. 
It is considered to be a fundamental problem of 3D computer vision, as establishing correspondences is a precondition for several downstream tasks such as structure from motion~\cite{heinly2015reconstructing,schoenberger2016sfm}, visual localisation~\cite{svarm2017city,sattler2017are,taira2018inloc} or simultaneous localization and mapping~\cite{strasdat2011double,sarlin2022lamar}. 
Despite the abundant work dedicated to image matching over the last thirty years, this topic remains an unsolved problem for challenging scenarios, such as when images are captured from strongly differing viewpoints~\cite{arnold2022mapfree} \cite{jin2020image}, present occlusions~\cite{sarlin2022lamar} or feature day-to-night changes~\cite{zhang2021reference}.

An inherent part of the problem is the difficulty to evaluate image matching methods. 
In practice, it is often tackled with a proxy task such as relative or absolute camera pose estimation, or 3D reconstruction.
\\
In this context, image matching performances were recently significantly improved with the advent of attention layers~\cite{vaswani2017attention}. 
Cross-attention layers are mainly responsible for this breakthrough~\cite{sarlin2020superglue} as they enable the local features of detected keypoints in both images to communicate and adjust with respect to each other.
Prior siamese architectures~\cite{yi2016lift,ono2018lf,dusmanu2019d2}, \cite{revaud2019r2d2}, \cite{germain2020s2dnet,germain2021neural} had so far prevented this type of communication.
Shortly after this breakthrough, a second significant milestone was reached~\cite{sun2021loftr} by combining the usage of attention layers~\cite{sarlin2020superglue} with the idea of having a detector-free method~\cite{rocco2018neighbourhood,rocco2020efficient,rocco2020ncnet,li2020dual}, \cite{zhou2021patch2pix,truong2021learning}.
In LoFTR~\cite{sun2021loftr}, low-resolution dense cross-attention layers are employed that allow semi-dense low-resolution features of the two images to communicate and adjust to each other.  
Such a method  is said to be detector-free, as it matches semi-dense local features instead of sparse sets of local features coming from detected keypoint locations~\cite{lowe1999object}.

Semi-dense Detector-Free (\adf{}) methods~\cite{chen2022aspanformer,giang2022topicfm,wang2022matchformer,sun2021loftr}, \cite{mao20223dg,tang2022quadtree}, such as LoFTR, are among the best performing image matching approaches in terms of pose estimation metrics.
However, to the best of our knowledge, the link between their ability to establish correspondences and the quality of the resulting estimated pose has thus far received little attention.
This paper is a first attempt to study this link.

\medskip

\noindent \textbf{Contributions - }
We start with proposing a novel Structured Attention-based image Matching architecture (\poda).
We evaluate \poda{} and 6 \adf{} methods on 3 datasets (MegaDepth - relative pose estimation and matching, HPatches - homography estimation and matching, ETH3D - matching).

We highlight a counter-intuitive result on two datasets (MegaDepth and HPatches): on the one hand \poda{} either outperforms or is on par with \adf{} methods in terms of pose/homography estimation metrics, but on the other hand \adf{} approaches are significantly better than \poda{} in terms of Matching Accuracy (MA). Here the MA is computed on all the semi-dense locations (of the source image) with available ground truth correspondent, which includes both textured and uniform regions. 

We propose to limit the computation of the matching accuracy to textured regions, and show that in this case \poda{} often surpasses \adf{} methods. Our findings highlight a strong correlation between the ability to establish accurate correspondences in textured regions and the accuracy of the resulting estimated pose/homography (see Figure~\ref{fig:teaser}).

\medskip

\noindent \textbf{Organization of the paper - }
Sec.~\ref{sec:related_work} discusses the related work. 
The proposed \poda{} architecture is introduced in Sec.~\ref{sec:method} and the experiments in Sec.~\ref{sec:exp}.

\section{Related work}\label{sec:related_work}

Since this paper focuses on the link between the ability of \adf{} methods to establish correspondences and the quality of the resulting estimated pose, we only present \adf{} methods in this literature review and refer the reader to \cite{edstedt2023dkm,zhu2023pmatch,ni2023pats} for a broader literature review.

To the best of our knowledge, LoFTR~\cite{sun2021loftr} was the first method to perform attention-based detector-free matching. 
A siamese CNN is first applied on the source/target image pair to extract fine dense features of resolution 1/2 and coarse dense features of resolution 1/8.  
The source and target coarse features are fed into a dense attention-based module, interleaving self-attention layers with cross-attention layers as proposed in~\cite{sarlin2020superglue}. 
To reduce the computational complexity of these dense attention layers, Linear Attention~\cite{katharopoulos2020transformers} is used instead of vanilla softmax attention~\cite{vaswani2017attention}. 
The resulting features are matched to obtain coarse correspondences.
Each coarse correspondence is then refined by cropping $5\!\times\!5$ windows into the fine features and applying another attention-based module. Thus for each location of the semi-dense (factor of 1/8) source grid, a correspondent is predicted. In practice, a Mutual Nearest Neighbor (MNN) step is applied at the end of the coarse matching stage to remove outliers. 

In~\cite{tang2022quadtree}, a QuadTree attention module is proposed to reduce the computational complexity of vanilla softmax attention from quadratic to linear while keeping its power, as opposed to Linear Attention~\cite{katharopoulos2020transformers} which was shown to underperform on local feature matching~\cite{germain2022visual}. 
The QuadTree attention module is used as a replacement for Linear Attention module in LoFTR. 
An architecture called ASpanFormer is introduced in~\cite{chen2022aspanformer} that employs the same refinement stage as LoFTR but a different coarse stage architecture. 
Instead of classical cross-attention layers, the coarse stage uses global-local cross-attention layers that have the ability to focus on regions around current potential correspondences. 
MatchFormer~\cite{wang2022matchformer} proposes an attention-based backbone, interleaving self and cross-attention layers, to progressively transform a tensor of size $2\!\times\!H\!\times\!W\!\times\!3$ into a tensor of size $2\!\times\!\frac{H}{16}\!\times\!\frac{W}{16}\!\times\!C$. Efficient attention layers such as Spatial Efficient Attention~\cite{wang2021pyramid,xie2021segformer} and Linear Attention~\cite{katharopoulos2020transformers} are employed. 
A feature pyramid network-like decoder~\cite{lin2017feature} is used to output a fine tensor of size $2\!\times\!\frac{H}{2}\!\times\!\frac{W}{2}\!\times\!C_f$ and a coarse tensor of size $2\!\times\!\frac{H}{8}\!\times\!\frac{W}{8}\!\times\!C_c$, and the matching is performed as in LoFTR.
TopicFM~\cite{giang2022topicfm} follows the same global architecture as LoFTR with a different coarse stage. 
Here, topic distributions (latent semantic instances) are inferred from coarse CNN features using cross-attention layers with topic embeddings. 
These topic distributions are used to augment the coarse CNN features with self and cross-attention layers.
In~\cite{mao20223dg}, a method called 3DG-STFM proposes to train the LoFTR architecture with a student-teacher method. 
The teacher is first trained on RGB-D image pairs. 
The teacher model then guides the student model to learn RGB-induced depth information.

\begin{figure*}[ht]
\centering
{\small
\begin{tabular}{cccc}
  \hspace{0.3cm} Source Image & \hspace{1.5cm} Target Image &\hspace{1.2cm} Average query map &\hspace{1.cm} Average latent map  \\
\end{tabular} 
}
    \includegraphics[width=0.95\textwidth,height=0.29\textwidth]{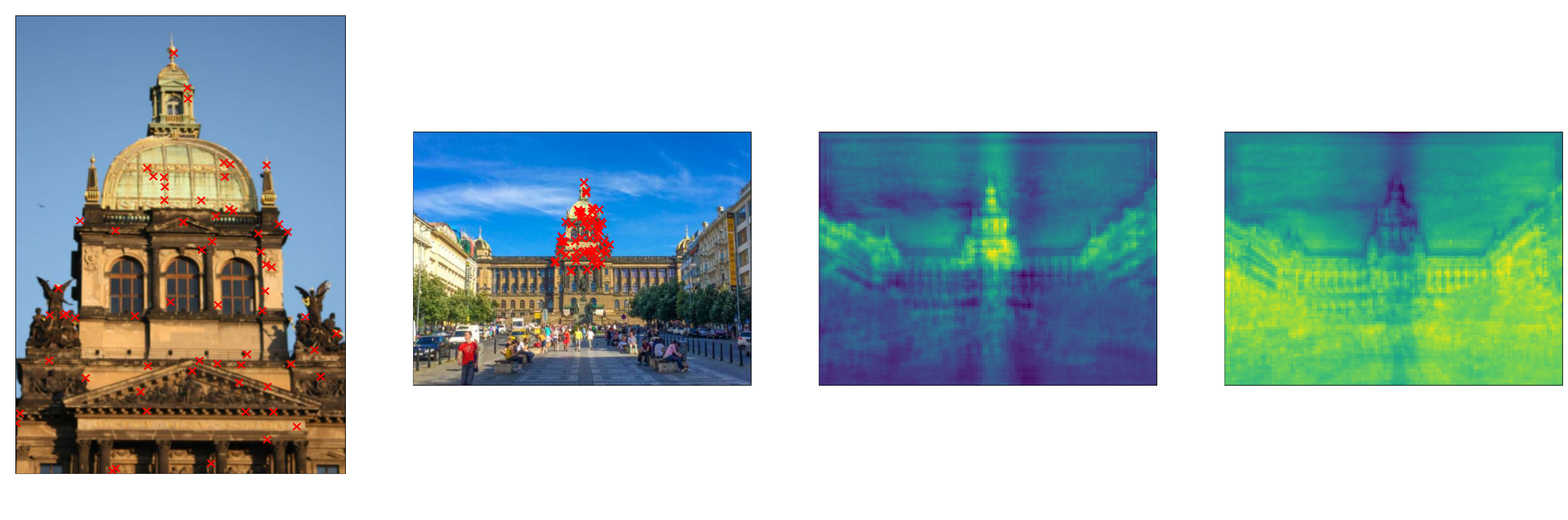}\vspace{-0.2cm}
    \caption{\textbf{Visualization of the learned latent vectors of \poda{}.} The average query map is obtained by averaging 64 correspondence maps of 64 query locations (\textcolor{red}{\textbf{Red}} crosses) while the average latent map is obtained by averaging the 128 correspondence maps of the 128 learned latent vectors. We observe that the average query map is mainly activated around the correspondents, whereas these regions are less activated in the average latent map.} 
\label{fig:latent_space}
\end{figure*}
\section{Structured attention-based image matching architecture}\label{sec:method}
In the following section, we present our novel Structured Attention-based image Matching architecture (\poda), whose architecture  is illustrated in Figure~\ref{fig:baseline}.

\subsection{Background and notations}

\noindent\textbf{Image matching - } Given a set of 2D query locations $\curl{\vp_{\text{s},i}}_{i=1...L}$ in a source image $\mI_\text{s}$, we seek to find their 2D correspondent locations $\curl{\hat{\vp}_{\text{t},i}}_{i=1...L}$  in a target image $\mI_\text{t}$:
\begin{align}
\curl{\hat{\vp}_{\text{t},i}}_{i=1...L}=\mathcal{M}\parr{\mI_{\text{s}},\mI_{\text{t}},\curl{\vp_{\text{s},i}}_{i=1...L}}.\label{eq:odfm}
\end{align}
Here $\mathcal{M}$ is the image matching method. In the case of semi-dense detector-free methods, the query locations $\curl{\vp_{\text{s},i}}_{i=1...L}$ are defined as the source grid locations using a stride of 8. As we will see, \poda{} is more flexible and can process any set of query locations. However, for a fair comparison, in the experiments all the methods (including \poda{}) will use source grid locations using a stride of 8 as query locations.

\medskip

\noindent\textbf{Vanilla softmax attention - } A cross-attention operation, with $H$ heads, between a $D$-dimensional query vector $\vx$ and a set of $D$-dimensional vectors $\curl{\vy_n}_{n=1...N}$ can be written as follows:
\begin{align}
&\sum_{h=1}^H \sum_{n=1}^N  \eqcolor{\overbrace{\black{s_{h,n}}}^{1{\times}1}} \eqcolor{\overbrace{\black{\mW_{\text{o},h}}}^{D\times D_H}}  \eqcolor{\overbrace{\black{\mW_{\text{v},h}}}^{D_H{\times}D}} \eqcolor{\overbrace{\black{\vy_n}}^{D\times1}},\\
\text{where } &\quad s_{h,n}=\frac{\exp(\eqcolor{\overbrace{\black{\vx\tr}}^{1\times D}} \eqcolor{\overbrace{\black{\mW_{\text{q},h}\tr}}^{D\times D_H}} \eqcolor{\overbrace{\black{\mW_{\text{k},h}^{\phantom{\top}}}}^{D_H \times D}}\eqcolor{\overbrace{\black{\vy_n\tr}}^{D\times 1}})}{\sum_{m=1}^N \exp(\vx\tr \mW_{\text{q},h}\tr \mW_{\text{k},h}\vy_m)}.\label{eq:coeff}
\end{align}

In practice, all the matrices $\mW_{\cdot,\cdot}$ are learned. 
Usually $D_H=\frac{D}{H}$. 
The output is a linear combination (with coefficients $s_{h,n}$) of the linearly transformed set of vectors $\curl{\vy_n}_{n=1...N}$. 
This operation allows to extract the relevant information in $\curl{\vy_n}_{n=1...N}$ from the point of view of the query $\vx$. 
The query $\vx$ is not an element of that linear combination. 
As a consequence, a residual connection is often added, usually followed by a layer normalization and a 2-layer MLP (with residual connection)~\cite{vaswani2017attention}.

In practice, a cross-attention layer between a set of query vectors $\curl{\vx_n}_{n=1...Q}$ and $\curl{\vy_n}_{n=1...N}$ is performed in parallel, which is often computationally demanding since, for each head, the coefficients are stored in a $Q\!\times\!N$ matrix.
A self-attention layer is a cross-attention layer where $\curl{\vy_n}_{n=1...N}=\curl{\vx_n}_{n=1...Q}$.

\subsection{Feature extraction stage }
Our method takes as input a source image $\mI_\text{s}$ ($H_\text{s}\!\times\!W_\text{s}\!\times3$), a target image $\mI_\text{t}$ ($H_\text{t}\!\times\!W_\text{t}\!\times3$) and a set of 2D query locations ${\curl{\vp_{\text{s},i}}_{i=1...L}}$. 
The first stage of \poda{} is a classical feature extraction stage. 
From the source image $\mI_\text{s}$ ($H_\text{s}\!\times\!W_\text{s}\!\times3$)  and the target image $\mI_\text{t}$ ($H_\text{t}\!\times\!W_\text{t}\!\times3$), dense visual source features $\mF_\text{s}$  ($\frac{H_\text{s}}{4}\!\times\!\frac{W_\text{s}}{4}\!\times128$) and target features $\mF_\text{t}$  ($\frac{H_\text{t}}{4}\!\times\!\frac{W_\text{t}}{4}\!\times128$)  are extracted using a siamese CNN backbone. 
The Positional Encodings (PE) of the source and target are computed, using an MLP~\cite{sarlin2020superglue}, and concatenated with the visual features of the source and target to obtain two tensors $\mH_\text{s}$ ($\frac{H_\text{s}}{4}\!\times\!\frac{W_\text{s}}{4}\!\times256$) and $\mH_\text{t}$ ($\frac{H_\text{t}}{4}\!\times\!\frac{W_\text{t}}{4}\!\times256$). 
For each 2D query point $\vp_{\text{s},i}$, a descriptor $\vh_{\text{s},i}$ of size 256 is extracted from $\mH_\text{s}$. In practice, we use integer query locations thus there is no need for interpolation here. 
Technical details concerning this stage are provided in the appendix.

\subsection{Latent space attention-based stage}
In order to allow for the descriptors ${\curl{\vh_{\text{s},i}}_{i=1...L}}$ (at training-time we use $L=1024$) to communicate and adjust with respect to $\mH_\text{t}$, we draw inspiration from Perceiver~\cite{jaegle2021perceiver,jaegle2022perceiver}, and consider a set of $N\!=\!M\!+\!L$ latent vectors, composed of $M$ learned latent vectors $\curl{\vm_i}_{i=1...M}$ and the $L$ descriptors $\curl{\vh_{\text{s},i}}_{i=1...L}$. 
These latent vectors $\curl{\curl{\vm_i}_{i=1...M},\curl{\vh_{\text{s},i}}_{i=1...L}}$ are used as queries in an \emph{input} cross-attention layer to extract the relevant information from $\mH_\text{t}$, and finally obtain an updated set of latent vectors $\curl{\curl{\vm^{(0)}_i}_{i=1...M},\curl{\vh^{(0)}_{\text{s},i}}_{i=1...L}}$.
On the one hand, the outputs $\curl{\vh^{(0)}_{\text{s},i}}_{i=1...L}$ contain the information relevant to find their respective correspondents within $\mH_\text{t}$. 
On the other hand, the outputs $\curl{\vm^{(0)}_i}_{i=1...M}$  extracted a general representation of $\mH_\text{t}$ since they are not aware of the query 2D locations.
In practice, we set $M$ to 128.
Afterward, a series of $S$ self-attention layers are applied to the latent vectors to get $\curl{\curl{\vm^{(S)}_i}_{i=1...M},\curl{\vh^{(S)}_{\text{s},i}}_{i=1...L}}$.
In these layers, all latent vectors can communicate and adjust with respect to each other. 
For instance, the set $\curl{\vm^{(s)}_i}_{i=1...M}$ can be used to disambiguate certain correspondences.
 Then,  $\mH_\text{t}$ is used as a query in an \emph{output} cross-attention layer to extract the relevant information from the latent vectors. 
 The resulting tensor is written $\mH^{\text{out}}_\text{t}$. 
 For each updated descriptor $\vh^{(S)}_{\text{s},i}$, a correspondence map $\mC_{\text{t},i}$ (of size $\frac{H_\text{t}}{4}\!\times\!\frac{W_\text{t}}{4}$) is obtained by computing the dot product between $\vh^{(S)}_{\text{s},i}$ and $\mH^{\text{out}}_\text{t}$. 
 Finally, for each 2D query location $\vp_{\text{s},i}$, the predicted 2D correspondent location $\hat{\vp}_{\text{t}
,i}$ is defined as the argmax of $\mC_{\text{t},i}$.

In Figure~\ref{fig:latent_space}, we propose a visualization of the learned latent vectors of \poda{}.
For visualization purposes, we used $L=64$. Thus the average query map is obtained by averaging the 64 correspondence maps of the 64 query locations, while the average latent map is obtained by averaging the 128 correspondence maps of the 128 learned latent vectors.
We observe that the average query map is mainly activated around the correspondents, whereas these regions are less activated in the average latent map.

\begin{figure*}
{\small
  \hspace{0.5cm} {Source Image}   \hspace{2cm} Target Image   \hspace{1.5cm} Visio-positional map  \hspace{1.5cm} Positional map  \\
}
\centering
    \includegraphics[width=0.95\textwidth,height=0.15\textwidth]{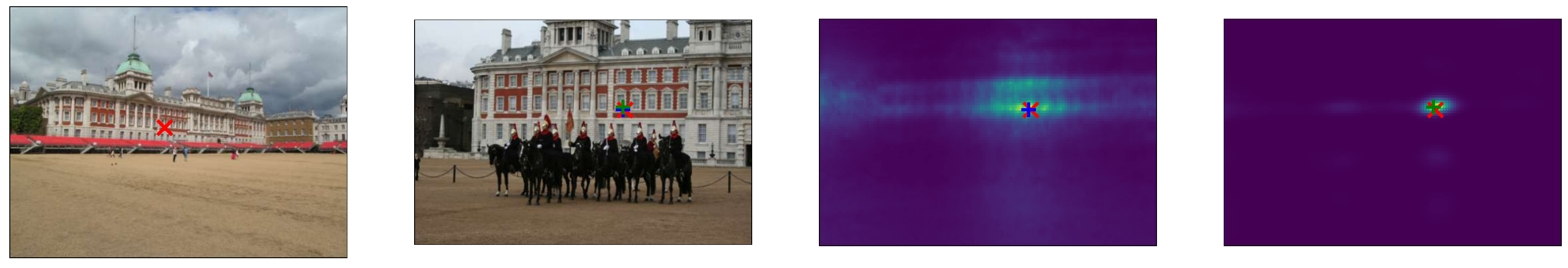}
    \includegraphics[width=0.95\textwidth,height=0.15\textwidth]{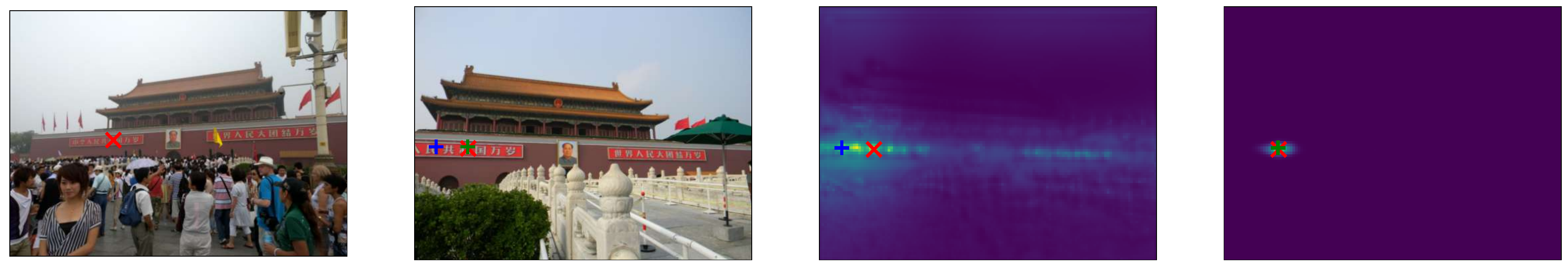}
    \caption{\textbf{Visualization - Structured attention.} The visio-positional and positional maps are computed before the output cross-attention. {\color{red}\textbf{Red}} crosses represent the ground-truth correspondences. {\color{blue}\textbf{Blue}} and \textcolor{green}{\textbf{Green}} crosses are the maxima of the visio-positional and positional maps, respectively. One can see that the visio-positional maps are highly multimodal (\emph{i.e.}, sensitive to repetitive structures) while the positional maps are almost unimodal.} 
\label{fig:corresp_maps}
\end{figure*}

\subsection{Structured attention}
At the end of the feature extraction stage, the upper half of each vector is visual features, while the lower half is positional encodings. 
In each attention layer (input cross-attention, self-attention and output cross-attention), we found it important to structure the linear transformation matrices $\mW_{\text{o},h}$ and  $\mW_{\text{v},h}$ to constrain the lower half of each output vector to only contain a linear transformation of the positional encodings:
\vspace{-0.15cm}
\begin{align}
\eqcolor{\overbrace{\black{\mW_{\text{o},h}}}^{D\times D_H}}=\bigg [{\begin{array}{ccc}
\multicolumn{3}{c}{\eqcolor{\overbrace{\black{\mW_{\text{o},h,\text{up}}}}^{\frac{D}{2}\times D_H}}}\\
\hline
\eqcolor{\underbrace{\black{\m 0}}_{\
\frac{D}{2}\times \frac{D_H}{2}}} &\! | \!& \eqcolor{\underbrace{\black{\mW_{\text{o},h,\text{low}}}}_{\frac{D}{2}\times \frac{D_H}{2}}} 
\end{array}}\bigg ].
\end{align}
The matrices $\mW_{\text{v},h}$ and the fully connected layers within the MLP (at the end of each attention module) are structured the exact same way. 
Consequently, throughout the network, the lower half of each latent vector only contains transformations of positional encodings, but no visual feature.

In Figure~\ref{fig:corresp_maps} we propose a visualization of two different correspondence maps \emph{before} the output cross-attention.

The first map is produced using the first 128 dimensions of the feature representation and contains high-level visio-positional features.
The second map is produced using the 128 last dimensions of the feature representation and contains only positional encodings.
By building these distinct correspondence maps we can observe that while the visio-positional representation is sensitive to repetitive structures, the purely positional representation tends to be only activated in the areas neighbouring the match.

\medskip

\subsection{Loss}
At training time, we use a cross-entropy (CE) loss function~\cite{germain2020s2dnet} on each correspondence map $\mC_{\text{t},i}$ in order to maximise its score at the ground truth location $\vp_{\text{t},i}$:
\begin{equation}
\text{CE}\parr{\mC_{\text{t},i},\vp_{\text{t},i}} = -\ln\parr{\frac{\exp\, \mC_{\text{t},i}\parr{\vp_{\text{t},i}}}{\sum_{\v q} \exp \,\mC_{\text{t},i}\parr{\v q}}}.\label{eq:loss}
\end{equation}

\subsection{Refinement}
The previously described architecture produces coarse correspondence maps of resolution 1/4. 
Thus the predicted correspondents ${\curl{\hat\vp_{\text{t},i}}_{i=1...L}}$ need to be refined. 
To do so, we simply use a second siamese CNN that outputs dense source and target features at full resolution. 
For each 2D query location $\vp_{\text{s},i}$, a correspondence map is computed on a window of size 11 centered around the coarse prediction.  
The predicted 2D correspondent location $\hat{\vp}_{\text{t},i}$ is defined as the argmax of this correspondence map.  
This refinement network is trained separately using the same cross-entropy loss eq.~\ref{eq:loss}. 
Technical details are provided in the appendix.

\begin{table*}
\centering
\footnotesize
\caption{Evaluation on MegaDepth1500~\cite{sarlin2020superglue}.
We report Matching Accuracy \eqref{eq:ap}
for several thresholds $\eta$,
computed on the set of all the semi-dense query locations (source grid with stride 8) with available ground truth correspondents (\ap),
and on a subset containing only query locations that are within a textured region of the source image (\apkp). Concerning the pose estimation metrics, we report the classical AUC at 5, 10 and 20 degrees.
The proposed \poda{} method outperforms \adf{} methods in terms of pose estimation, while \adf{} methods are significantly better in terms of \ap. However, when uniform regions are ignored (\apkp), \poda{} often
surpasses \adf{} methods. These results highlight a strong
correlation between the ability to establish precise
correspondences in textured regions and the accuracy of
the resulting estimated pose.
}
\vspace{-0.2cm}
{\scriptsize
\begin{tabular}{@{\hspace{1mm}}c@{\hspace{2mm}}l@{\hspace{2mm}}c
@{\hspace{2mm}}c@{\hspace{2mm}}c@{\hspace{2mm}}c@{\hspace{2mm}}c@{\hspace{2mm}}c@{\hspace{2mm}}c@{\hspace{6mm}}c@{\hspace{2mm}}c@{\hspace{2mm}}c@{\hspace{2mm}}c@{\hspace{2mm}}c@{\hspace{2mm}}c@{\hspace{2mm}}c@{\hspace{6mm}}c@{\hspace{2mm}}c@{\hspace{3mm}}c@{\hspace{2mm}}c@{\hspace{2mm}}}
\specialrule{.1em}{1em}{0em} 
&\multirow{3}{*}{ Method}  
&\multicolumn{6}{@{\hspace{0mm}}c@{\hspace{0mm}}}{\multirow{2}{*}{Matching Accuracy (\ap) \eqref{eq:ap} $\uparrow$}} && \multicolumn{6}{@{\hspace{0mm}}c@{\hspace{0mm}}}{{Matching Accuracy}} && \multicolumn{3}{@{\hspace{0mm}}c@{\hspace{0mm}}}{{Pose estimation on}}& \\ 
&&&&&&& && \multicolumn{6}{@{\hspace{0mm}}c@{\hspace{0mm}}}{{on textured regions (\apkp) \eqref{eq:ap} $\uparrow$}} && \multicolumn{3}{@{\hspace{0mm}}c@{\hspace{0mm}}}{1/8 grid (AUC) $\uparrow$} \\[0.5ex] \cline{3-8} \cline{10-15} \cline{17-19}
\\[-2.5ex]
                        &&    $\eta$=$1$ & $\eta$=$2$ & $\eta$=$3$ & $\eta$=$5$ & $\eta$=$10$ & $\eta$=$20$ && $\eta$=$1$ & $\eta$=$2$ & $\eta$=$3$ & $\eta$=$5$ & $\eta$=$10$  & $\eta$=$20$&& $@5^\text{o}$ &$@10^\text{o}$ &$@20^\text{o}$  \\ 
         \hline 
         \\[-2.25ex]
&LoFTR \cite{sun2021loftr}   & 49.7 & 73.2 & 81.6 & 87.4 & 90.5 & 91.8 &&
55.3 & 75.2 & 81.6 & 86.9 & 89.9 & 91.9 &&
52.8 & 69.2 & 82.0\\
&MatchFormer \cite{wang2022matchformer} & 51.1 & 73.4 & 81.0 & 86.9 & 89.5 & 90.9 &&
56.5 & 75.6 & 81.8 & 87.1 & 89.6 & 90.9&&
52.9 & 69.7 & 82.0\\
&TopicFM \cite{giang2022topicfm} & 51.4 & 75.4 & 83.7 & 89.9 & 92.9 & 93.5 && 
59.8 & 77.6 & 84.8 & 90.4 & 92.9 & 93.7 &&
54.1 & 70.1 & 81.6\\
&3DG-STFM\cite{mao20223dg} & 51.6 & 73.7 & 80.7 & 86.4 & 89.0 & 90.7 && 
57.0 & 75.8 & 81.8 & 86.8 & 88.8 & 90.5 &&
52.6 & 68.5 & 80.0\\
&ASpanFormer \cite{chen2022aspanformer}& \textbf{52.0} & \textbf{76.2} & \textbf{84.5} & \textbf{90.7} & \textbf{93.7} & \textbf{94.8}  &&
\underline{62.2} & \underline{80.3} & \underline{85.9} & \textbf{91.0} & \textbf{93.7} & \underline{94.7} &&
\underline{55.3} & \underline{71.5} & \underline{83.1}\\
&LoFTR+QuadTree \cite{tang2022quadtree} & \underline{51.6} & \underline{75.9} & \underline{84.1} & \underline{90.2} & \underline{93.1} & \underline{94.0} &&
61.7 & 79.9 & 85.5 & 90.5 & 93.3 & 94.1 &&
54.6 & 70.5 & 82.2\\[0.5ex]
  \noalign{\global\arrayrulewidth=0.2mm}
  \arrayrulecolor{gray!30}\hline
       \\[-2.25ex]
&\textbf{\poda{}  (ours)}  & 48.5 & 70.4 & 78.0 & 83.0 & 85.4 & 86.4 &&
\textbf{67.9} & \textbf{83.8 } & \textbf{87.3} & \underline{90.6} & \underline{93.6} & \textbf{95.2}&&
\textbf{55.8} & \textbf{72.8} & \textbf{84.2}\\[-2ex]     
\arrayrulecolor{black}
\specialrule{.1em}{1em}{0em} 
 \end{tabular}
}

\label{table:megadepth_tab}
 \end{table*}

\begin{table*}
\centering
\footnotesize
\caption{Evaluation on HPatches~\cite{balntas2017hpatches}. We report Matching Accuracy \eqref{eq:ap}
for several thresholds $\eta$,
computed on the set of all the semi-dense query locations (source grid with stride 8) with available ground truth correspondents (\ap),
and on a subset containing only query locations that are within a textured region of the source image (\apkp). 
Concerning the homography estimation metrics, we report the classical AUC at 3, 5 and 10 pixels.
The proposed \poda{} method is on par with \adf{} methods in terms of homography estimation, while \adf{} methods are significantly better in terms of \ap. However, when uniform regions are ignored (\apkp), \poda{}
matches \adf{} performances. These results highlight a strong
correlation between the ability to establish precise
correspondences in textured regions and the accuracy of
the resulting estimated homography.}

{\scriptsize
\begin{tabular}{@{\hspace{1mm}}c@{\hspace{2mm}}l@{\hspace{2mm}}c
@{\hspace{4mm}}c@{\hspace{4mm}}c@{\hspace{2mm}}c@{\hspace{6mm}}c@{\hspace{5mm}}c@{\hspace{4mm}}c@{\hspace{0mm}}c@{\hspace{6mm}}c@{\hspace{4mm}}c@{\hspace{3mm}}c@{\hspace{1mm}}c@{\hspace{2mm}}}
\specialrule{.1em}{1em}{0em} 
&\multirow{3}{*}{ Method}  
&\multicolumn{3}{@{\hspace{0mm}}c@{\hspace{0mm}}}{Matching Accuracy} && \multicolumn{3}{@{\hspace{0mm}}c@{\hspace{0mm}}}{{Matching Accuracy on}} && \multicolumn{3}{@{\hspace{0mm}}c@{\hspace{0mm}}}{{Homography estimation}}& \\ 
&& \multicolumn{3}{@{\hspace{0mm}}c@{\hspace{0mm}}}{{(\ap) \eqref{eq:ap} $\uparrow$}} && \multicolumn{3}{@{\hspace{0mm}}c@{\hspace{0mm}}}{{textured regions (\apkp) \eqref{eq:ap} $\uparrow$}} && \multicolumn{3}{@{\hspace{0mm}}c@{\hspace{0mm}}}{(AUC) $\uparrow$} \\[0.5ex] \cline{3-5} \cline{7-9} \cline{11-13}
\\[-2.5ex]
                        &&  $\eta$=$3$  & $\eta$=$5$ & $\eta$=$10$ & & \text{ } $\eta$=$3$ \text{ } &  \text{ } $\eta$=$5$ \text{ } & $\eta$=$10$ && $@3 \text{px}$ & $@5 \text{px}$ & $@10 \text{px}$  \\ 
         \hline 
         \\[-2.25ex]
&LoFTR \cite{sun2021loftr}  & 66.8 & 74.3& 77.3 &&
67.6 & 75.3 & 78.4&&
65.9 & 75.6 & 84.6   \\
&MatchFormer \cite{wang2022matchformer} & 66.2 & 74.9 & 78.2&&
67.7 & 75.8 & 79.1 &&
65.0&73.1&81.2\\
&TopicFM \cite{giang2022topicfm}   &  \underline{72.7} & \underline{85.0} & \underline{87.5} &&
\textbf{74.0} & \underline{86.0} & \underline{88.5}&&
\underline{67.3}& \textbf{77.0} & \underline{85.7}\\
&3DG-STFM\cite{mao20223dg} & 64.9 & 75.1 & 78.2 &&
66.2 & 74.3 & 77.6 &&
64.7&73.1&81.0\\
&ASpanFormer \cite{chen2022aspanformer} & \textbf{76.2} & \textbf{86.2} & \textbf{88.7} &&
\underline{73.9} & 85.8 & 88.4 &&
\textbf{67.4}&\underline{76.9}&85.6\\
&LoFTR+QuadTree \cite{tang2022quadtree} & 70.2 & 83.1 & 85.9 &&
73.5 & 84.3 & 86.9 &&
67.1&76.1&85.3\\[0.5ex]
  \noalign{\global\arrayrulewidth=0.2mm}
  \arrayrulecolor{gray!30}\hline
       \\[-2.25ex]
&\textbf{\poda{} (ours)} &
62.4 & 70.9 & 74.2&&
73.4 & \textbf{86.6} & \textbf{89.3}&&
67.1 &\underline{76.9}& \textbf{85.9} \\[-2ex]     
\arrayrulecolor{black}
\specialrule{.1em}{1em}{0em} 
 \end{tabular}
}
\label{table:hpatches}
 \end{table*}

\begin{table*}
\centering
\footnotesize
\caption{Evaluation on ETH3D~\cite{schops2019bad} for different frame interval sampling rates $r$. We report Matching Accuracy \eqref{eq:ap}
for several thresholds $\eta$,
computed on the set of all the semi-dense query locations (source grid with stride 8) with available ground truth correspondents (\ap). For ETH3D, ground truth correspondents are based on structure from motion tracks. Consequently, the \ap{} already ignores untextured regions of the source images which explains why \poda{} is able to outperform \adf{} methods.}
\vspace{-0.2cm}
{\scriptsize
\begin{tabular}{@{\hspace{1mm}}c@{\hspace{2mm}}l@{\hspace{2mm}}c
@{\hspace{2mm}}c@{\hspace{2mm}}c@{\hspace{2mm}}c@{\hspace{2mm}}c@{\hspace{2mm}}c@{\hspace{6mm}}c@{\hspace{2mm}}c@{\hspace{2mm}}c@{\hspace{2mm}}c@{\hspace{2mm}}c@{\hspace{2mm}}c@{\hspace{6mm}}c@{\hspace{2mm}}c@{\hspace{2mm}}c@{\hspace{2mm}}c@{\hspace{2mm}}c@{\hspace{2mm}}c@{\hspace{2mm}}}
\specialrule{.1em}{1em}{0em} 
&\multirow{3}{*}{ Method}  
&\multicolumn{17}{@{\hspace{0mm}}c@{\hspace{0mm}}}{Matching Accuracy (\ap) \eqref{eq:ap} $\uparrow$ } \\[0.5ex]
\cline{3-19}
\\[-2ex]
&& \multicolumn{5}{@{\hspace{0mm}}c@{\hspace{0mm}}}{$r=3$} &&  
 \multicolumn{5}{@{\hspace{0mm}}c@{\hspace{0mm}}}{$r=7$} &&
 \multicolumn{5}{@{\hspace{0mm}}c@{\hspace{0mm}}}{$r=15$} \\[0.5ex]
 \cline{3-7} \cline{9-13} \cline{15-19}
\\[-2.5ex]
                        &&    $\eta$=$1$ & $\eta$=$2$ & $\eta$=$3$& $\eta$=$5$ & $\eta$=$10$   && $\eta$=$1$ & $\eta$=$2$ & $\eta$=$3$ &  $\eta$=$5$ & $\eta$=$10$ &&  $\eta$=$1$ & $\eta$=$2$ & $\eta$=$3$& $\eta$=$5$ & $\eta$=$10$   \\ 
         \hline 
         \\[-2.25ex]
&LoFTR \cite{sun2021loftr} &44.8 & 76.5 & 88.4 & 97.0&99.4&&
39.7&73.1& 87.6 &95.9&98.5&&
33.3&66.2& 84.8 &92.5&96.3\\
&MatchFormer \cite{wang2022matchformer}&45.5&77.1& 89.2 &97.2&99.7&&
40.4&73.8& 87.8 &96.6&99.0&&
34.2&66.7& 84.9 &93.5&97.0\\
&TopicFM \cite{giang2022topicfm} & 45.1 &76.9& 89.0 &97.2&99.6&&
39.9&73.5& 87.9 &96.4&99.0&&
33.8&66.4& 85.0 &92.8&96.5\\
&3DG-STFM\cite{mao20223dg} & 43.9&76.3& 88.0 &96.9&99.3&&
39.3&72.7& 87.4 &95.5&98.3&&
32.4& 65.7&  84.7 &92.0&96.0\\
&ASpanFormer \cite{chen2022aspanformer}&45.8&\underline{77.6}& \underline{89.6} & \underline{97.8}&\textbf{99.8}&&
40.6&73.8& 88.1 &96.8&99.0&&
34.3&66.8&  85.3 &93.9&97.3\\
&LoFTR+QuadTree \cite{tang2022quadtree} & \underline{45.9}&77.5 & 89.5 &97.8&99.7&&
\underline{40.8}&\underline{74.0}& \underline{88.3} &\underline{97.0}&\underline{99.2}&&
\underline{34.5}&\underline{66.8}& \underline{85.4} &\underline{94.0}&\underline{97.3}\\[0.5ex]
  \noalign{\global\arrayrulewidth=0.2mm}
  \arrayrulecolor{gray!30}\hline
      \\[-2.25ex]
&\textbf{\poda{} (ours)} & \textbf{53.4} & \textbf{79.9} & \textbf{91.5} & \textbf{98.0} & \textbf{99.8} &&
\textbf{48.6} & \textbf{78.6} & \textbf{91.7} & \textbf{98.2} & \textbf{99.4} &&
\textbf{40.1} & \textbf{70.2} & \textbf{87.8} & \textbf{95.4} & \textbf{97.8} \\[-2ex]     
\arrayrulecolor{black}
\specialrule{.1em}{1em}{0em} 
 \end{tabular}
}
\label{table:eth3D}
 \end{table*}
 
\section{Experiments}\label{sec:exp}

In these experiments, we focus on evaluating six \adf{} networks (LoFTR~\cite{sun2021loftr}, QuadTree~\cite{tang2022quadtree}, ASpanFormer~\cite{chen2022aspanformer}, 3DG-STFM~\cite{mao20223dg}, MatchFormer~\cite{wang2022matchformer}, TopicFM~\cite{giang2022topicfm}) and our proposed architecture \poda{}.
In all the following Tables, best and second best results are respectively bold and underlined.

\medskip

\noindent\textbf{Implementations - } For each network, we employ the code and weights (trained on MegaDepth) made available by the authors. Concerning \poda{}, we train it similarly to the  \adf{} networks on MegaDepth~\cite{li2018megadepth}, during 100 hours, on four GeForce GTX 1080 Ti (11GB) GPUs (see the appendix for details). Our code will be made available.

\medskip
\vspace{-0.025cm}
 
\noindent\textbf{Query locations - } Concerning \adf{} methods, the query locations are defined as the source grid locations using a stride of 8. Thus, in order to be able to compare the performances of \poda{} against such methods, we use the exact same query locations. To do so  (recall that \poda{} was trained with $L=1024$ query locations), we simply shuffle the source grid locations (with stride 8) and feed them to \poda{} by batch of size 1024 (the CNN features are cached making the processing of each minibatch very efficient).

\medskip
\vspace{-0.025cm}

\noindent\textbf{Evaluation criteria - } Regarding the pose and homography estimation metrics, we use the classical AUC metrics for each dataset. Thus in each Table, the pose/homography results we obtained for \adf{} methods (we re-evaluated each method) are the same as the results published in the respective papers. Concerning \poda{}, the correspondences are classically filtered (similarly to what \adf{} methods do) before the pose/homography estimator using a simple Mutual Nearest Neighbor with a threshold of 5 pixels.

In order to evaluate the ability of each method to establish correspondences, we consider the
Matching Accuracy (MA) as in \cite{truong2020glu}, 
\emph{i.e.}, the average on all images of the ratio of correct matches, for different pixel error thresholds ($\eta$): \vspace{-0.05cm}
\begin{align}
\text{MA}\parr{\eta}=\frac{1}{K}\sum_{k=1}^K\frac{ \sum_{i=1}^{L_k} \croch{\normtwo{\hat{\vp}_{\text{t}_k,i}-\vp^{\text{GT}}_{\text{t}_k,i}}<\eta}}{ L_k},  \label{eq:ap}
\end{align}
where $K$ is the number of pairs of images, $L_k$ is the number of ground truth correspondences available for the image pair $\#k$ and $\croch{\cdot}$ is the Iverson bracket. We refer to this metric as \ap{} and not "Percentage of Correct Keypoints" because we found this term misleading in cases where the underlying ground truth correspondences are not based on keypoints, as it is the case in MegaDepth and HPatches.

We also propose to introduce the Matching Accuracy in textured regions (\apkp) that consists in ignoring, in eq.~(\ref{eq:ap}), ground truth correspondences whose query location is within a low-contrast region of the source image, \emph{i.e.}, an almost uniform region of the source image.

Note that \poda{}'s MNN (and the MNN of \adf{} methods) is only used for the pose/homography estimation, \emph{i.e.}, it is not used to compute matching accuracies, otherwise the set of correspondences would not be the same for each method.


\subsection{Evaluation on MegaDepth} 
We consider the MegaDepth1500~\cite{sarlin2020superglue} benchmark. We use the exact same settings as those used by \adf{} methods, such as an image resolution of 1200. The results are provided in Table~\ref{table:megadepth_tab}. We report Matching Accuracy \eqref{eq:ap}, for several thresholds $\eta$,
computed on the set of all the semi-dense query locations (source grid with stride 8) with available ground truth correspondents (\ap). The ground truth correspondents are obtained using the available depth maps and camera poses. Consequently, many query locations located in untextured regions have a ground truth correspondent. Thus, we also report the matching accuracy computed only on query locations that are within a textured region of the source image (\apkp). Concerning the pose estimation metrics, we report the classical AUC at 5, 10 and 20 degrees.
The proposed \poda{} method outperforms \adf{} methods in terms of pose estimation, while \adf{} methods are significantly better in terms of \ap. However, when uniform regions are ignored (\apkp), \poda{} often
surpasses \adf{} methods. These results highlight a strong
correlation between the ability to establish precise
correspondences in textured regions and the accuracy of
the resulting estimated pose.

\begin{figure*}
\centering
{\footnotesize
\begin{tabular}{@{\hspace{2mm}}c@{\hspace{12mm}}c}
{\poda{} (ours)}  &  
{LoFTR+QuadTree} \cite{tang2022quadtree}  \\[1ex]
\ap@2=60.6    \apkp@2=\textbf{84.0}    errR=\textbf{2.1'}    errT=\textbf{5.4'} &
\ap@2=\textbf{62.8}    \apkp@2=64.0    errR=5.5'    errT=22.4'\\
  \multicolumn{2}{c}{\includegraphics[width=\textwidth,height=0.26\textwidth]{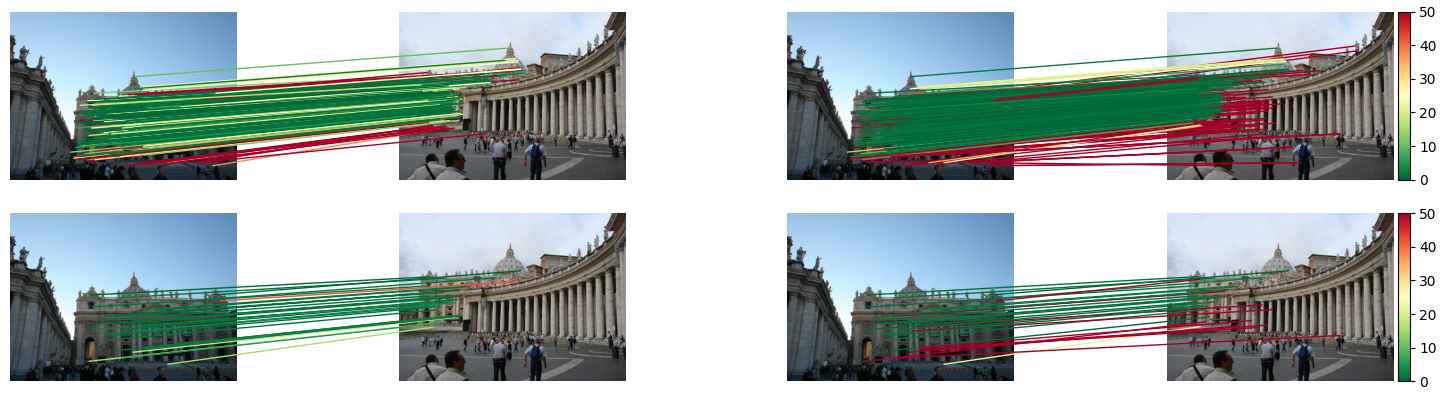}}
 \\[-0.5ex]
  \hline \\[-1.5ex]
 \hspace{-0mm} \ap@2=62.3 \apkp@2=\textbf{81.1} errR=\textbf{1.9'} errT=\textbf{5.5'}  & 
  \hspace{10mm} \ap@2=\textbf{66.2} \apkp@2=67.5 errR=2.1' errT=9.4' \\
  \multicolumn{2}{c}{\includegraphics[width=\textwidth,height=0.26\textwidth]{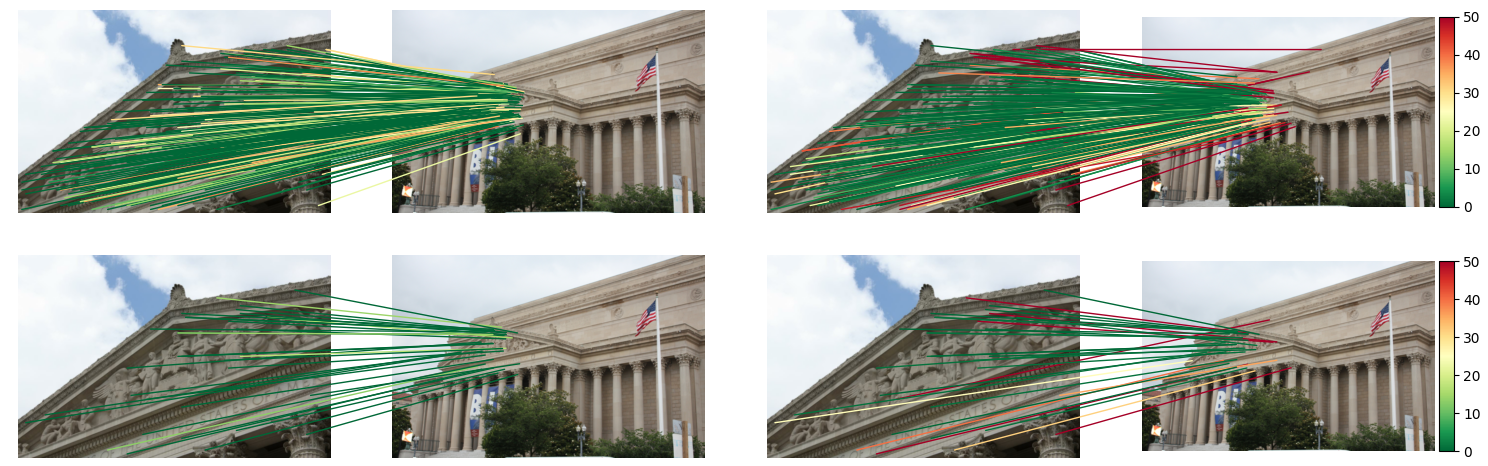}}
    \\ \hline \\ [-1.5ex]
      \hspace{-0mm} \ap@2=79.0 \apkp@2=\textbf{83.1} errR=\textbf{1.5'} errT=\textbf{3.9'}  & 
  \hspace{10mm} \ap@2=\textbf{79.9} \apkp@2=78.7 errR=2.0' errT=6.4' \\
  \multicolumn{2}{c}{\includegraphics[width=\textwidth,height=0.26\textwidth]{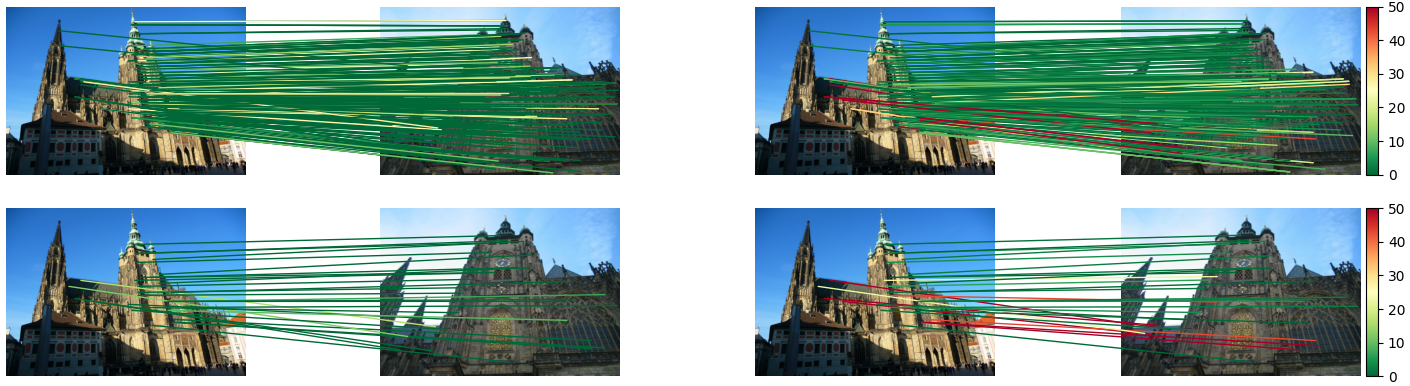}}  \\ \hline \\ [-1.5ex]
  \hspace{-0mm} \ap@5=69.1 \apkp@5=\textbf{78.0} errR=\textbf{3.8'} errT=\textbf{15.4'}  & 
  \hspace{10mm} \ap@5=\textbf{70.7} \apkp@5=72.1 errR=4.9' errT=19.6' \\
  \multicolumn{2}{c}{\includegraphics[width=\textwidth,height=0.26\textwidth]{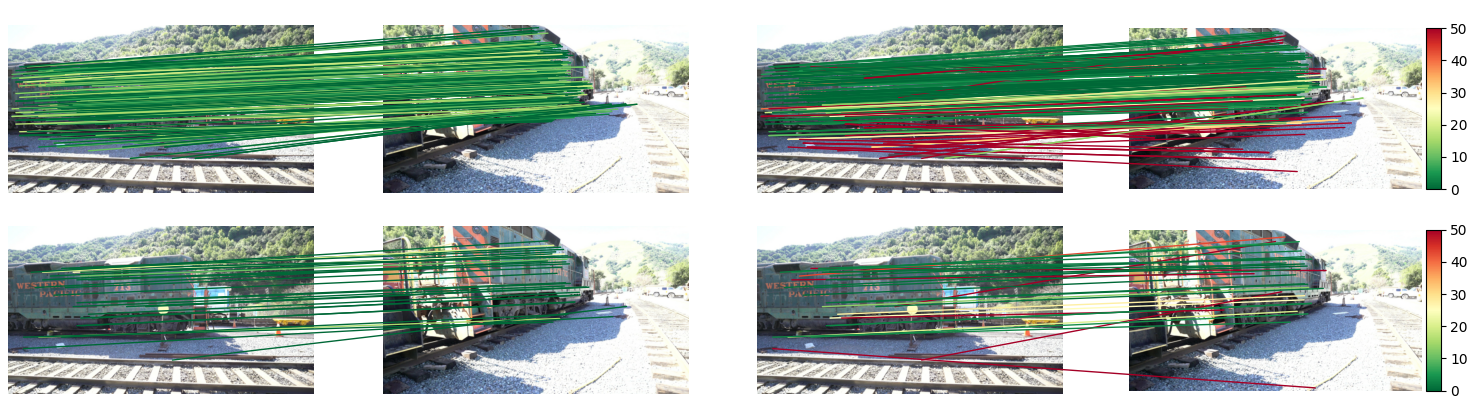}}
   \end{tabular}
}
   \caption{\textbf{Qualitative results on MegaDepth1500.} For each image pair: (top row) Visualization of established correspondences used to compute the \ap{}, (bottom row) Visualization of established correspondences used to compute the \apkp{}. Line colors indicate the distance in pixels \wrt the ground truth correspondent.} \vspace{-0.2cm}
\label{fig:qualitative}
\end{figure*}

In Figure~\ref{fig:qualitative}, we report qualitative results that visually illustrate the previous findings.

\subsection{Evaluation on HPatches} 
We evaluate the different architectures on HPatches~\cite{balntas2017hpatches} (see Table~\ref{table:hpatches}). 
We use the exact same settings as those used by \adf{} methods. We report Matching Accuracy \eqref{eq:ap}
for several thresholds $\eta$,
computed on the set of all the semi-dense query locations (source grid with stride 8) with available ground truth correspondents (\ap). The ground truth correspondents are obtained using the available homography matrices. Consequently, many query locations located in untextured regions have a ground truth correspondent. Thus, we also report the matching accuracy computed only on query locations that are within a textured region of the source image (\apkp).
Concerning the homography estimation metrics, we report the classical AUC at 3, 5 and 10 pixels.
The proposed \poda{} method is on par with \adf{} methods in terms of homography estimation, while \adf{} methods are significantly better in terms of \ap. However, when uniform regions are ignored (\apkp), \poda{}
matches \adf{} performances. These results highlight a strong
correlation between the ability to establish precise
correspondences in textured regions and the accuracy of
the resulting estimated homography.

\begin{figure*}[ht]
\centering
{\small
\begin{tabular}{@{\hspace{4mm}}c@{\hspace{12mm}}c}
\hspace{1.5cm} 
{\poda{} (ours)}  &  
\hspace{.3cm} 
{LoFTR+QuadTree} \cite{tang2022quadtree}  \\[1ex]
  \hspace{-0mm} \ap@5=69.7 \apkp@5=\textbf{88.6} AUC@5=\textbf{77.4}  & 
  \hspace{-5mm} \ap@5=\textbf{72.1} \apkp@5=76.6 AUC@5=76.1\\
  \multicolumn{2}{c}{\includegraphics[width=0.9\textwidth,height=0.25\textwidth]{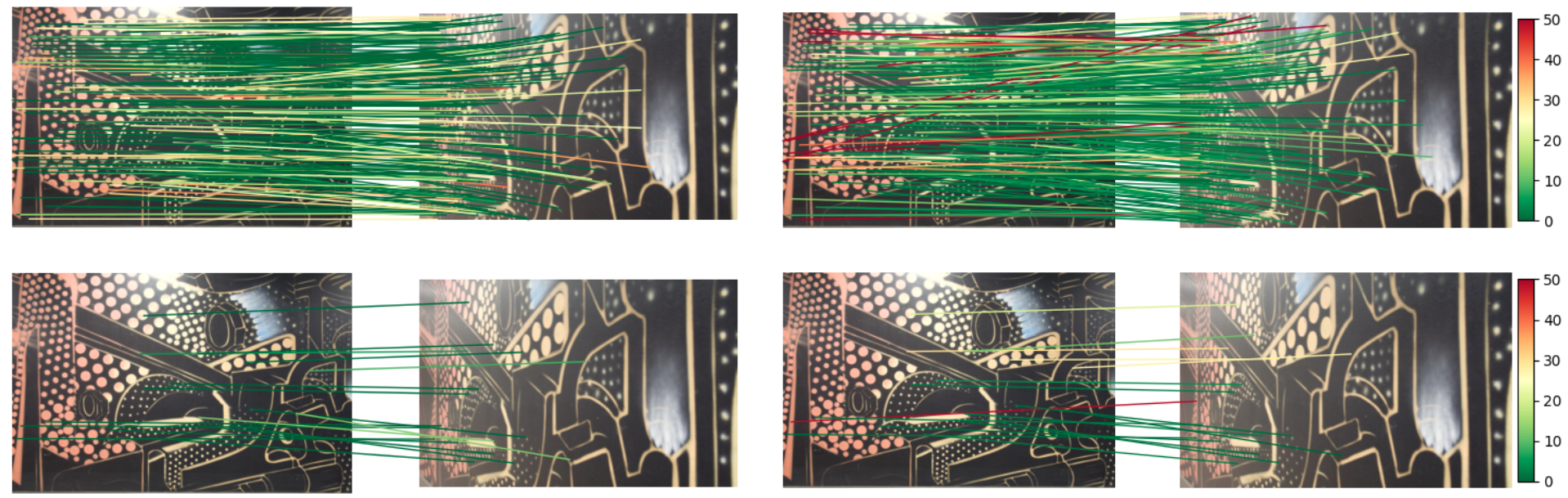}}
  \\ \hline \\ [-1.5ex]
  \hspace{-0mm} \ap@5=74.4 \apkp@5=\textbf{86.2} AUC@5=\textbf{81.5}  & 
  \hspace{-5mm} \ap@5=\textbf{77.1} \apkp@5=42.8 AUC@5=75.9\\
  \multicolumn{2}{c}{\includegraphics[width=0.9\textwidth,height=0.28\textwidth]{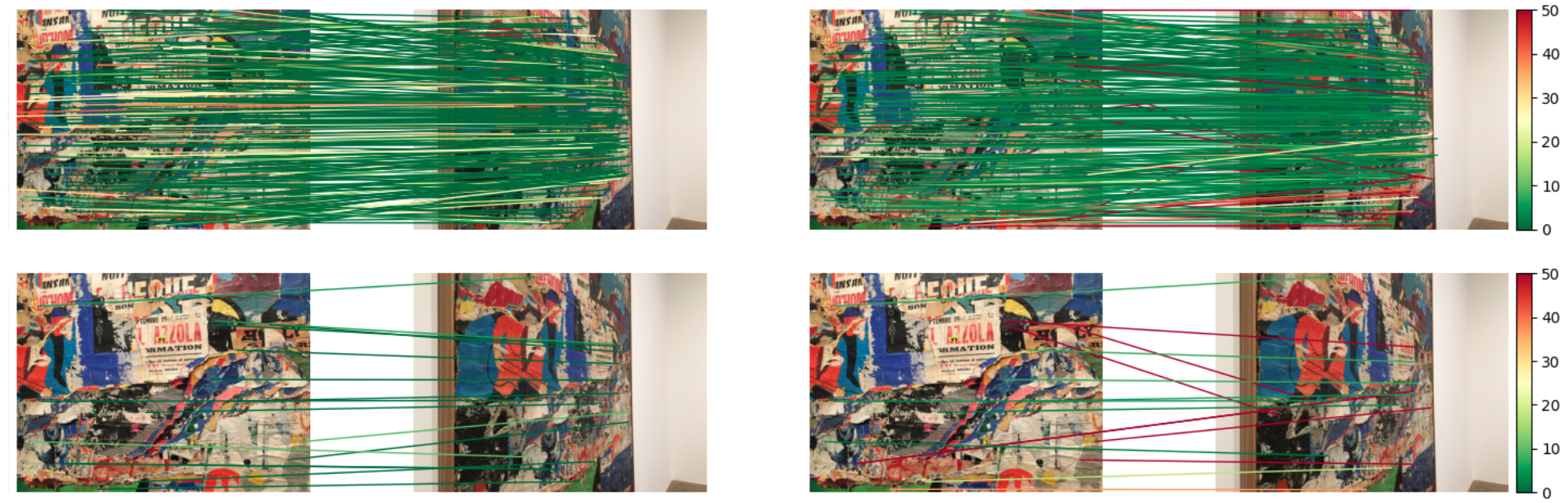}}
%
\end{tabular}
}
   \caption{\textbf{Qualitative results on HPatches.} Line colors indicate the distance in pixels to the ground truth correspondent.
\vspace{0.3cm}} 
\label{fig:hpatches}
\end{figure*}

In Figure~\ref{fig:hpatches}, we report qualitative results that visually illustrate the previous findings. 


\subsection{Evaluation on ETH3D} 
We evaluate the different networks on several sequences of the ETH3D dataset~\cite{schops2019bad} as proposed in \cite{truong2020glu}.
Different frame interval sampling rates $r$ are considered. As the rate $r$  increases, the overlap between the image pairs reduces, hence making the matching problem more difficult. 
The results are provided in Table ~\ref{table:eth3D}.
We report Matching Accuracy \eqref{eq:ap} for several thresholds $\eta$,
computed on the set of all the semi-dense query locations (source grid with stride 8) with available ground truth correspondents (\ap). For ETH3D, ground truth correspondents are based on structure from motion tracks. Consequently, the \ap{} ignores untextured regions of the source images which explains why \poda{} is able to outperform \adf{} methods in terms of \ap{}.

\begin{figure*}[ht]
\centering
{\small
\begin{tabular}{@{\hspace{4mm}}c@{\hspace{12mm}}c}
\hspace{2.5cm} 
{\poda{} (ours)}  &  
\hspace{2.5cm} 
{LoFTR+QuadTree} \cite{tang2022quadtree}  \\[1ex]
 \hspace{-25mm}  \apkp@2=\textbf{75.0} & \hspace{-35mm} \apkp@2=68.8\\
   \multicolumn{2}{c}{\includegraphics[width=\textwidth]{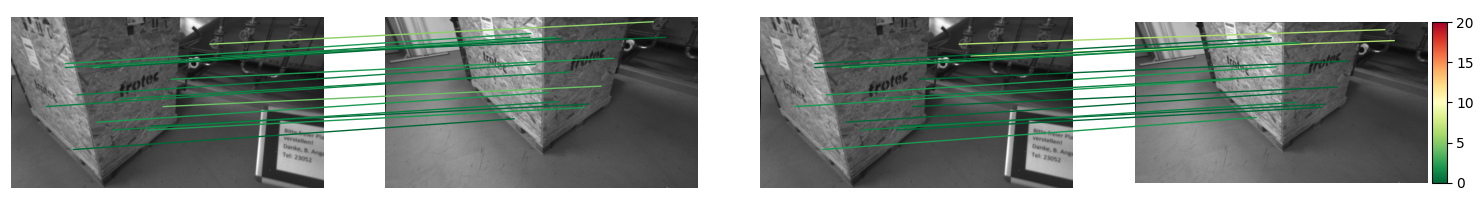}}
 \\ \hline \\ [-1.5ex]
  \hspace{-25mm}  \apkp@2=\textbf{65.4} & \hspace{-35mm} \apkp@2=61.5\\
   \multicolumn{2}{c}{\includegraphics[width=\textwidth]{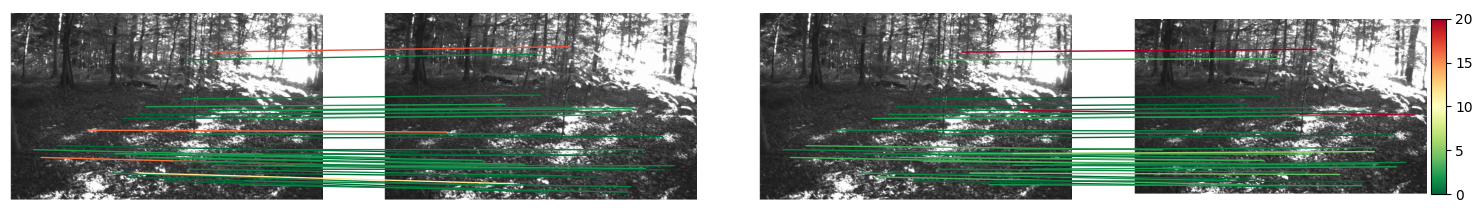}}
  \\ \hline \\ [-1.5ex]
 \hspace{-25mm}  \apkp@2=\textbf{80.0} & \hspace{-35mm} \apkp@2=68.0\\
 \multicolumn{2}{c}{\includegraphics[width=\textwidth]{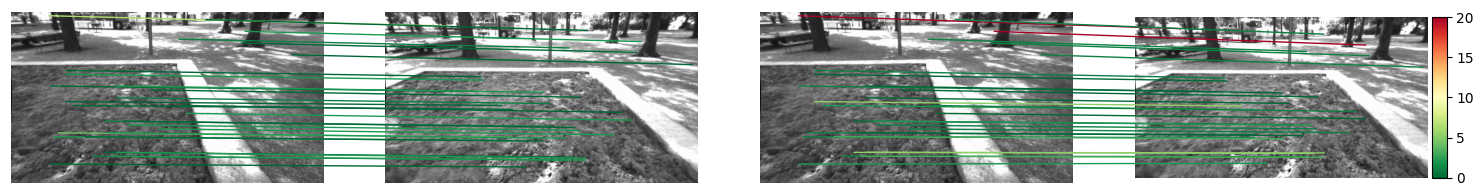}}
 \\ \hline \\ [-1.5ex]
 \hspace{-25mm}  \apkp@2=\textbf{87.2} & \hspace{-35mm} \apkp@2=69.2\\
  \multicolumn{2}{c}{\includegraphics[width=\textwidth]{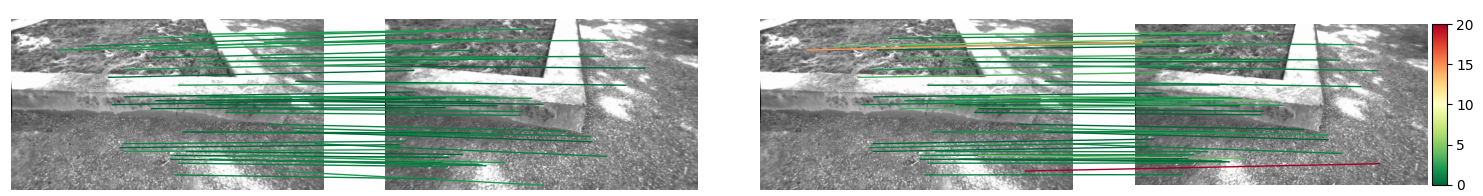}}
 \\ \hline \\ [-1.5ex]

\end{tabular}
}
   \caption{\textbf{Qualitative results on ETH3D.} Line colors indicate the distance in pixels to the ground truth correspondent.} \vspace{-0.2cm}
\label{fig:eth3D}
\end{figure*}

In Figure~\ref{fig:eth3D}, we report qualitative results that illustrate the accuracy of the proposed SAM method.

\subsection{Ablation study}

\begin{table}[t]
\centering
\caption{Ablation study of proposed \poda{} method (MegaDepth validation set).}
{\scriptsize
\begin{tabular}{@{\hspace{0.5mm}}l@{\hspace{2mm}}c
@{\hspace{2mm}}c@{\hspace{2mm}}c@{\hspace{2mm}}c@{\hspace{2mm}}c@{\hspace{2mm}}c@{\hspace{0.5mm}}}
\specialrule{.1em}{1em}{0em} 
\multirow{2}{*}{ Method}   &\multicolumn{6}{@{\hspace{0mm}}c@{\hspace{0mm}}}{\apkp $\uparrow$} \\ \cline{2-7}
                        & $\eta$=$1$ & $\eta$=$2$ & $\eta$=$5$ & $\eta$=$10$ & $\eta$=$20$ &  $\eta$=$50$ \\ 
         \hline
         \\[-2.25ex]
{\scriptsize Siamese CNN}                                & 0.029 & 0.112 & 0.436 & 0.581 & 0.622 & 0.687 \\
{\scriptsize \text{ }+ Input CA and SA (x16)}       & 0.137 & 0.321 & 0.671 & 0.734 & 0.767 & 0.822 \\
{\scriptsize \text{ }+ Learned LV and output CA}         & 0.132 & 0.462 & 0.823 & 0.871 & 0.898 & 0.935 \\
{\scriptsize \text{ }+ PE concatenated}                  & 0.121 & 0.419 & 0.796 & 0.868 & 0.902 & 0.939 \\
{\scriptsize \text{ }+ Structured AM }                   & \underline{0.140} & \underline{0.487} & \underline{0.857} & \underline{0.902} & \textbf{0.922} & \textbf{0.947} \\
{\scriptsize \text{ }+ Refinement (\textbf{Full model}) }& \textbf{0.673}& \textbf{0.791} & \textbf{0.870} & \textbf{0.902} & \underline{0.921} & \underline{0.946} \\[-2ex]  
\specialrule{.1em}{1em}{0em} 
\vspace{-0.1cm}
 \end{tabular}
 }
 \label{table:ablation}
\end{table}

In Table \ref{table:ablation}, we propose an ablation study of \poda{} to evaluate the impact of each part of our architecture.
This study is performed on MegaDepth validation scenes.
Starting from a standard siamese CNN, we show that a significant gain in performance can simply be obtained with the input cross-attention layer (here the PE is added and not concatenated) and self-attention layers.
We then add learned latent vectors (LV) in the latent space and use an output cross-attention which again significantly improves the performance.
Concatenating the positional encoding information instead of adding it to the visual features reduces the matching accuracy at $\eta=2$ and $\eta=5$. However, combining it with structured attention leads to a significant improvement in terms of \ap{}.
Finally, as expected, the refinement step improves the matching accuracy for small pixel error thresholds.

\section{Conclusion\label{sec:conclusion}}

We proposed a novel Structured Attention-based image Matching architecture (\poda). The flexibility of this novel architecture allowed us to fairly compare it against \adf{} methods, \emph{i.e.}, in all the experiments we used the same query locations (source grid with stride
8). The experiments highlighted a counter-intuitive result on two datasets (MegaDepth and HPatches): on the one hand \poda{} either outperforms or is on par with \adf{} methods in terms of pose/homography estimation metrics, but on the other hand \adf{} approaches are significantly better than \poda{} in terms of Matching Accuracy (\ap{}). Here the \ap{} is computed on all the semi-dense locations of the source image with available ground truth correspondent, which includes both textured and uniform regions. 
We proposed to limit the computation of the matching accuracy to textured regions, and showed that in this case \poda{} often surpasses \adf{} methods. These findings highlighted a strong correlation between the ability to establish precise correspondences in textured regions and the accuracy of the resulting estimated pose/homography. We also evaluated the aforementioned methods on ETH3D which confirmed, on a third dataset, that \poda{} has a strong ability to establish correspondences in textures regions. We finally performed an ablation study of \poda{} to demonstrate that each part of the architecture is important to obtain such a strong matching capacity.

\section*{Acknowledgments}
This project has received funding from the french minist\`ere de l'Enseignement sup\'erieur, de la Recherche et de l'Innovation.
This work was granted access to the HPC resources of IDRIS under the allocation 2022-AD011012858 made by GENCI.

\bibliographystyle{apalike}
{\small
\bibliography{biblio}}

\section*{\uppercase{Appendix}}

\paragraph{\poda{} architecture.}

Each attention layer in \poda{} is a classical series of softmax attention, normalization, MLP and normalization layers with residual connections, except for the "output" cross-attention where we removed the MLP, the last normalization and the residual connection. 

We empirically choose a combination of 16 self-attention layers and 128 vectors in the learned latent space. 
We found this combination to be a good trade-off between performance and computational cost.

\paragraph{Backbone description.}
We use a modified ResNet-18 as our feature extraction backbone. 
We extract the feature map after the last ResBlock in which we removed the last ReLU.
The features are reduced to ${\frac{1}{4}}^{th}$ of the image resolution ($3 \times H \times W$) by applying a stride of 2 in the first and third layer.
The feature representation built by the backbone is of size $128 \times \frac{1}{4} H \times \frac{1}{4} W$.

\paragraph{Refinement description.}
The previous CNN backbone is used for refinement. 
We simply added a FPN (Feature Pyramid Network) module to extract the feature maps after the first convolutional layer and each ResNet layer.
It leaves us with a feature map of size $128 \times H \times W$ on which the refinement stage is performed. 

\paragraph{Structured attention.}
In our implementation, $D=256$, we use 8 heads, thus $D_H = 32$. 

\begin{figure}
\centering  
  \includegraphics[width=0.45\linewidth]{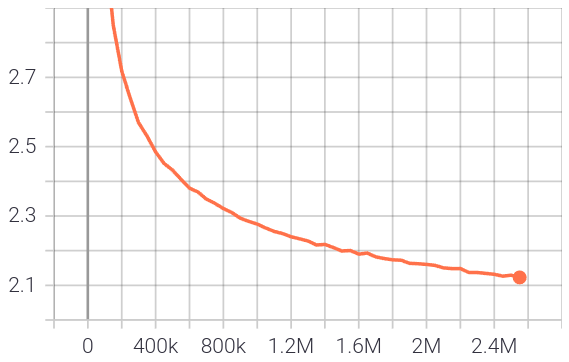}
  \includegraphics[width=0.45\linewidth]{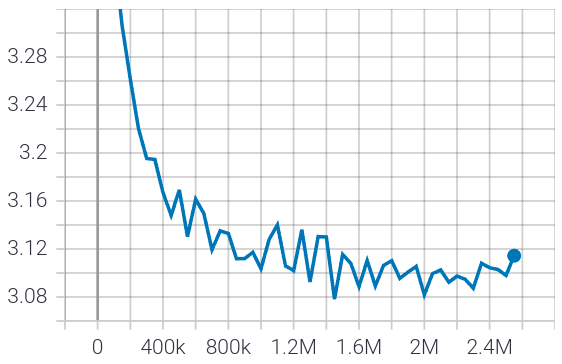}
  \caption{\textbf{Training (left) and Validation (right) loss for refinement ResNet-18+FPN.} x-axis represent the number of mini-batches seen by the model and y-axis the cross-entropy value.}
\label{fig:loss_backbone}
\end{figure}

\paragraph{Training details.}
The first part of the model to be trained is the refinement model (CNN+FPN).
It is trained during 50 hours on 4 Nvidia GTX1080 Ti GPUs using Adam optimizer, a constant learning rate of $10^{-3}$ and a batch size of 1 (Figure~\ref{fig:loss_backbone}).
We minimize the cross-entropy of the full resolution correspondence maps produced using the FPN.

\begin{figure}
\centering
\includegraphics[width=0.45\linewidth]{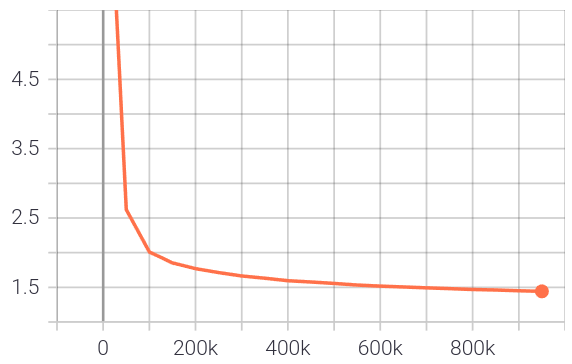}
  \includegraphics[width=0.45\linewidth]{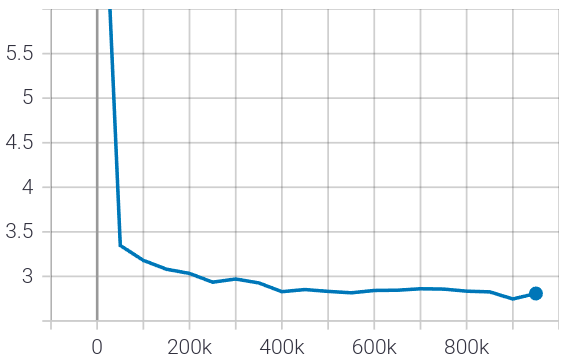}
  \caption{\textbf{Training (left) and Validation (right) loss for \poda{}.} x-axis represent the number of mini-batches seen by the model and y-axis the cross-entropy value.}
\label{fig:loss_POD}
\end{figure}

The CNN backbone of \poda{} is initialized with the weights of the previously trained network. We train \poda{} for 100 hours. The same setup with 4 Nvidia GTX1080 Ti GPUs, Adam optimizer and a batch size of 1 is used. Concerning the learning rate schedule, we use a linear warm-up of 5000 steps (from $0$ to $10^{-4}$) and then an exponential decay (towards $10^{-5}$)  (Figure~\ref{fig:loss_POD}). 
We minimize the cross-entropy of the $\frac{1}{4}$ resolution correspondence maps produced by \poda{}.

\end{document}